\documentclass[10pt,twocolumn,letterpaper]{article}

\usepackage{cvpr}      

\usepackage{algorithm}
\usepackage{algorithmic}
\usepackage{paralist}
\usepackage{tabularx}
\newcolumntype{C}{>{\centering\arraybackslash}X}
\usepackage[table]{xcolor}
\usepackage{newfloat}
\usepackage{listings}
\usepackage{amsmath} 
\usepackage{arydshln}
\usepackage{multirow}
\usepackage{bbding}
\usepackage{booktabs} 
\usepackage{float}
\usepackage{colortbl}
\usepackage{amssymb}
\usepackage{subcaption}
\definecolor{cvprblue}{rgb}{0.21,0.49,0.74}
\usepackage[pagebackref,breaklinks,colorlinks,allcolors=cvprblue]{hyperref}

\def\paperID{3071} 
\def\confName{CVPR}
\def\confYear{2026}

\title{Bridging Pixels and Words: Mask-Aware Local Semantic Fusion for Multimodal Media Verification}

\author{Zizhao Chen, ~Ping Wei\textsuperscript{\rm}\thanks{Corresponding author.}, ~~Ziyang Ren, ~Huan Li, ~Xiangru Yin\\
State Key Laboratory of Human-Machine Hybrid Augmented Intelligence,\\
Institute of Artificial Intelligence and Robotics, Xi'an Jiaotong University \\
{\tt\small bigd\_chen@stu.xjtu.edu.cn,~pingwei@xjtu.edu.cn}}

\begin{document}
\maketitle
\begin{abstract}
As multimodal misinformation becomes more sophisticated, its detection and grounding are crucial. However, current multimodal verification methods, relying on passive holistic fusion, struggle with sophisticated misinformation. Due to 'feature dilution,' global alignments tend to average out subtle local semantic inconsistencies, effectively masking the very conflicts they are designed to find. We introduce MaLSF (Mask-aware Local Semantic Fusion), a novel framework that shifts the paradigm to active, bidirectional verification, mimicking human cognitive cross-referencing. MaLSF utilizes mask-label pairs as semantic anchors to bridge pixels and words. Its core mechanism features two innovations: 1) a Bidirectional Cross-modal Verification (BCV) module that acts as an interrogator, using parallel query streams (Text-as-Query and Image-as-Query) to explicitly pinpoint conflicts; and 2) a Hierarchical Semantic Aggregation (HSA) module that intelligently aggregates these multi-granularity conflict signals for task-specific reasoning. In addition, to extract fine-grained mask-label pairs, we introduce a set of diverse mask-label pair extraction parsers. MaLSF achieves state-of-the-art performance on both the DGM4 and multimodal fake news detection tasks. Extensive ablation studies and visualization results further verify its effectiveness and interpretability.

\end{abstract}
\section{Introduction}

\begin{figure}[t]
  \centering
  \includegraphics[width=\linewidth]{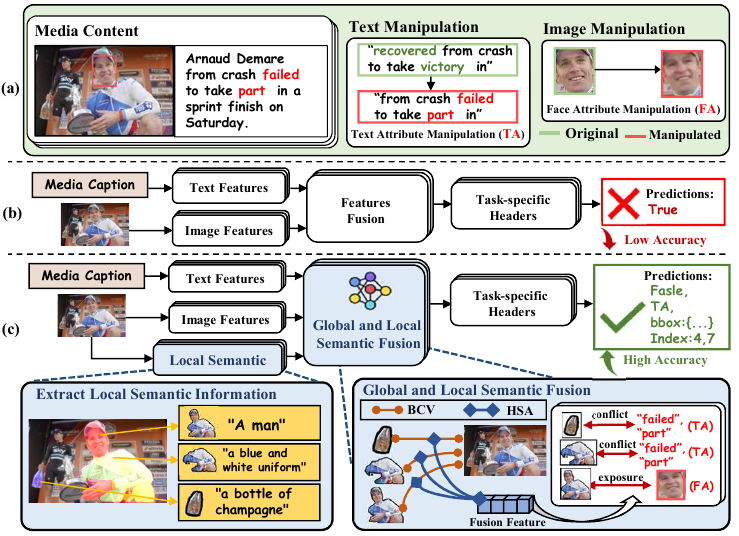}
  \caption{Overview of the misinformation and our framework. (a) A manipulated media content. (b) Traditional methods. (c) Our MaLSF framework. We utilize the BCV and HSA modules to perform explicit verification and aggregation of local semantics.}
  \label{fig1}
  \vspace{-0.5cm}
\end{figure}

The proliferation of advanced generative models \cite{cui2024instastyle,li2023pluralistic,cao2025avatargo} has ushered in an era of highly realistic multimodal deepfakes \cite{raza2023multimodaltrace,yin2024fine,qureshi2024deepfake}. Beyond simple factual mismatches, the most insidious forms of misinformation lie in subtle, local semantic inconsistencies \cite{shao2023detecting}. Take Fig. \ref{fig1}(a) as an example. It shows an image of an athlete celebrating with champagne, yet it is paired with a caption claiming he ``failed to take part in a sprint finish". While the global context (athlete, sprint) appears aligned, a critical, local semantic conflict exists between the visual evidence (``champagne") and the textual claim (``failed").  These cross-modal deceptions, characterized by complex manipulation strategies, are often difficult to detect and may thus exert widespread societal impacts~\cite{west2021misinformation,swire2020public}.

Faced with this challenge, current approaches \cite{liu2025modality,wang2023cross,10.1609/aaai.v37i4.25670,zhou2023multimodal,chen2022cross,shao2023detecting,shao2024detecting,liu2024fka,li2024towards,liu2024unified,wang2024exploiting} to multimodal verification are mainly built on a paradigm of ``passive" holistic fusion. As illustrated in Fig. \ref{fig1}(b), these methods encode the entire image and the entire text into high-dimensional vectors, which are then fused via attention or concatenation to produce a single, unified representation. The fundamental drawback in this paradigm is feature dilution. A semantically devastating, one-word change (e.g., ``recovered" to ``failed") may result in an almost imperceptible perturbation in the global text feature vector. When this ``globally-averaged" feature is fused with the visual representation, the subtle yet fatal conflict signal is averaged out by the overwhelming evidence of global semantic alignment. In our example, the global text vector would be barely changed by ``failed", and the ``champagne" visual feature would be averaged with the rest of the image, completely masking the conflict.

In contrast, human cognition does not perform passive fusion. When presented with such media, humans engage in a deliberate, ``active" verification process \cite{torres2018epistemology}. We cross-reference information by actively interrogating the modalities against each other. Upon reading ``failed", a human actively queries the image for visual evidence of ``failure." Conversely, upon seeing the ``champagne" in the image, the human queries the text for semantic correlates of ``victory." A conflict is flagged when this human bidirectional verification fails. This query-driven, bidirectional process is the key to identifying sophisticated, local-level misinformation, and this mechanism is absent in current models.

To computationally apply this human-centric strategy, we propose \textbf{MaLSF} (\textbf{M}ask-\textbf{a}ware \textbf{L}ocal \textbf{S}emantic \textbf{F}usion), a novel framework that shifts the paradigm from passive fusion to active, bidirectional verification. Our framework is built upon three key innovations that mirror this cognitive process. First, we introduce ``Mask-Label Pairs" as anchors, utilizing ``Mask-Label Pairs" \cite{deng2021transvg,liang2023open,zhang2023simple} as the foundational ``semantic anchors" that bridge pixel regions with their corresponding textual descriptions, enabling fine-grained, local-level reasoning. Second, we propose ``Bidirectional Cross-modal Verification (BCV)'' as the core engine of our framework. Instead of passively blending features, the BCV module acts as an ``interrogator" by implementing two parallel, query-driven verification streams (Text-as-Query and Image-as-Query) to pinpoint semantic conflicts explicitly. Finally, the ``Hierarchical Semantic Aggregation (HSA)'' module acts as the ``reasoning engine," intelligently aggregating the multi-granularity conflict signals discovered by the BCV through two stages, namely Multi-Label Shallow Fusion and Multi-Label Deep Fusion. It further encodes task-specific features in a hierarchically decoupled manner to support different verification branches.
This fusion-to-decoupling design enables the model to learn task-specific representations while maintaining semantic coherence.

To validate the comprehensive capabilities of our framework, we apply it to two key tasks in media integrity: Detecting and Grounding Multimodal Media Manipulation (DGM4)~\cite{shao2023detecting} and Multimodal Fake News Detection (MFND)~\cite{wang2018eann}. The DGM4 task is a complex, multifaceted challenge, requiring not only binary classification but also multi-label manipulation-type classification and precise grounding (localization) of forged content. The MFND task represents the core challenge of global authenticity verification (binary classification). We treat both as subtasks of a unified multimodal verification problem. We hypothesize that our fine-grained, query-driven architecture, designed to capture subtle local conflicts, provides a more robust foundation for both the complex, multi-output DGM4 task and the global-judgment MFND task.

As demonstrated in Fig. \ref{fig1}(c), MaLSF successfully identifies the subtle ``champagne-vs-failed" conflict that eludes traditional methods. Our experiments confirm that our framework achieves state-of-the-art performance on both tasks. Our contributions are summarized as follows:

\noindent $\bullet$ MaLSF shifts multimodal verification from passive fusion to an active, bidirectional verification process, mimicking human cognitive cross-referencing.

\noindent $\bullet$ We introduce ``Mask-Label Pairs" as semantic anchors and design the novel BCV and HSA modules to computationally realize this query-driven verification mechanism.

\noindent $\bullet$ We conduct comprehensive experiments on three datasets for the DGM4 task and MFND task. The experimental results show that the MaLSF model not only achieves state-of-the-art performance on both tasks but is also effective for different parsers.

\section{Related Work}

\subsection{Multimodal Fake News Detection}
Multimodal fake news detection (MFND) methods have explored diverse cross-modal interaction paradigms. Early works like EANN \cite{wang2018eann} introduced event discriminators to enhance feature learning, while MVAE \cite{khattar2019mvae} employed variational autoencoders for cross-modal reconstruction. MCAN \cite{wu2021multimodal} later leveraged stacked co-attention layers to fuse textual and visual features. To address modality reliability, CAFE \cite{chen2022cross} quantified cross-modal ambiguity via KL divergence to adaptively adjust feature weights, while LIIMR \cite{singhal2022leveraging} dynamically prioritized modalities based on confidence. More recently, COOLANT \cite{wang2023cross} aligns features via a cross-modal contrastive learning framework and adds a consistency learning task to optimize alignment accuracy. MIMoE-FND \cite{liu2025modality} uses a hierarchical hybrid expert architecture with an interactive mechanism for fine-grained multimodal reasoning. However, these methods primarily use coarse, global-level fusion (e.g., attention pooling, MoE) and lack explicit modeling of fine-grained semantic correspondences. This limitation becomes evident when handling subtle manipulations requiring localized evidence.

\begin{figure*}[tbp]
 \centering
 \includegraphics[width=\linewidth]{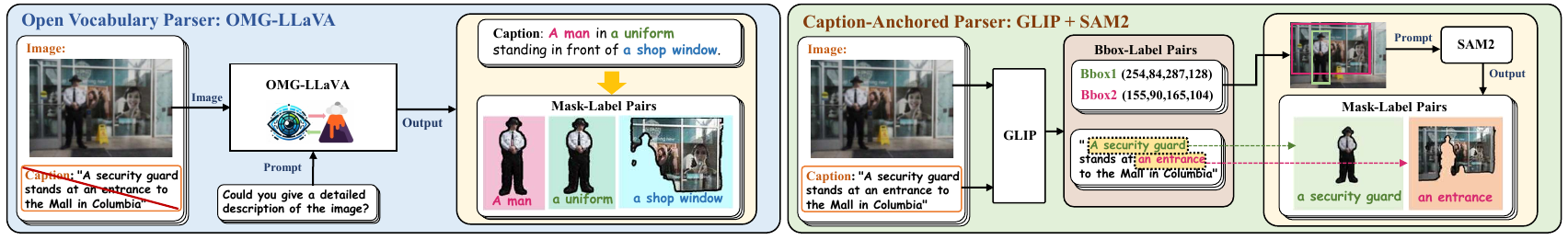}
\caption{Parsers for extracting mask-label pairs. Open Vocabulary Parser generates mask-label pairs end-to-end. Caption-Anchored Parser utilizes GLIP \cite{li2022grounded} to obtain labels and corresponding bboxes, and then SAM2 \cite{ravi2024sam} to obtain refined masks.}
\label{fig: parsers}
\vspace{-0.3cm}
\end{figure*}

\subsection{Multimodal Media Manipulation Detecting and Grounding}
To better localize media manipulation, Shao et al. \cite{shao2023detecting,shao2024detecting} proposed the Detecting and Grounding Multimodal Media Manipulation (DGM4) task and dataset. This task requires not only authenticity detection but also grounding manipulated content (image boxes, text tokens), demanding deeper multimodal reasoning. The pioneering works, HAMMER \cite{shao2023detecting} and HAMMER++ \cite{shao2024detecting}, use deep/shallow reasoning and contrastive learning for accurate detection and localization. UFAFormer \cite{liu2024unified} integrates image, text, and frequency-domain features (using DWT and self-attention) for joint modeling and accurate localization of cross-modal forgery. Wang et al. \cite{wang2024exploiting} proposed a Transformer framework using a two-branch cross-attention mechanism to fuse features, combined with an implicit manipulation query module and a decoupled classifier. Despite this progress, these approaches rely on global similarity or text-level comparisons, lacking explicit localization of conflicting semantics. Consequently, fine-grained alignment between visual regions and text claims remains underexplored. In this paper, we address these limitations by establishing “mask-label pairs” as the basic verification unit and realizing bidirectional cross-modal verification.

\vspace{-0.1cm}

\begin{figure*}[ht]
  \centering
  \includegraphics[width=\linewidth]{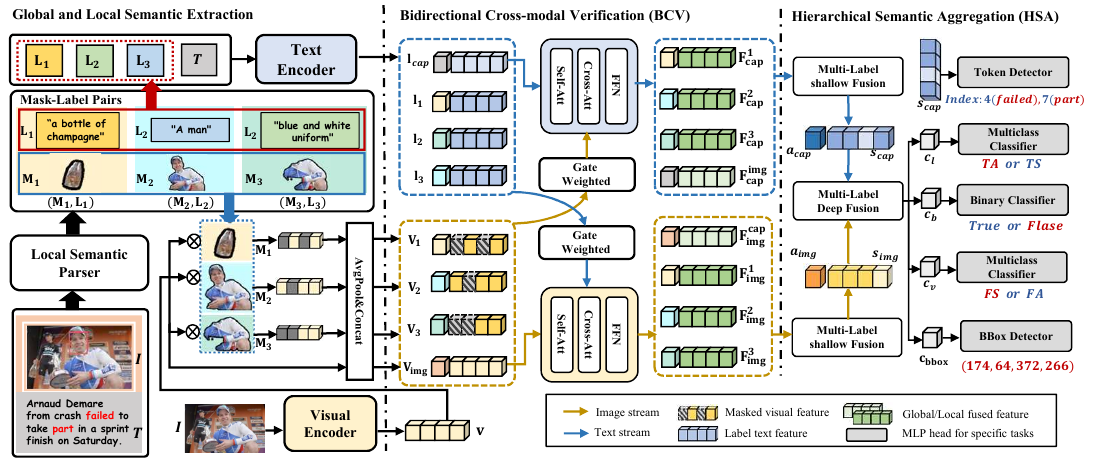}
  \caption{Overall architecture of MaLSF. In the figure, we take N=3 mask-label pairs as an example. The mask-label pairs combined with the original text and image are encoded as text features $\mathbf{l}_{\text{cap}},\mathbf{l}_{1},\mathbf{l}_{2},\mathbf{l}_{3}$ and image features $\mathbf{V}_{\text{img}},\mathbf{V}_{1},\mathbf{V}_{2},\mathbf{V}_{3}$ respectively. These features are subsequently encoded by BCV as multi-granularity validation features $\mathbf{F}_{\text{cap}}^{\text{img}}, \mathbf{F}^{\text{cap}}_{\text{img}},\mathbf{F}_{\text{cap}}^{i},\mathbf{F}^{i}_{\text{img}},i=1,2,3$.The HSA further aggregates the validation features to finally get the features $s_{\text{cap}},c_l,c_v,c_b,c_{\text{bbox}}$ used for different tasks. TA, TS, FA, and FS are different text and face manipulation types, respectively.}
  \label{fig2}
  \vspace{-0.3cm}
\end{figure*}

\section{Methods}

In this section, we will introduce our proposed MaLSF framework in detail. First, we will introduce our parser for obtaining mask-label pairs. Next, we will discuss the details of the modules in MaLSF. As shown in Fig. \ref{fig2}, the detection and grounding of media content will be divided into three stages: global and local semantic extraction, bidirectional cross-modal verification, and hierarchical semantic aggregation. Next, we will present our method in detail.

\subsection{Parsers for Extracting Mask-Label Pairs}
\label{pipeline}



To obtain sufficiently effective mask-label pairs as our basic unit, we propose two parsers, as illustrated in Fig. \ref{fig: parsers}. The Open Vocabulary Parser operates as a single-stage extraction model. Leveraging OMG-LLaVA \cite{zhang2024omg}, a large language model with advanced pixel-level visual understanding and reasoning capabilities, this parser processes each image to generate a descriptive caption and the corresponding mask-label pairs for objects mentioned within the caption. In contrast, the Caption-Anchored Parser employs a two-stage approach. First, the open-vocabulary object detector GLIP \cite{li2022grounded} processes the image alongside its original caption to identify and localize objects referenced in the caption, outputting their bounding boxes. Subsequently, these bounding boxes serve as prompts for SAM2 \cite{ravi2024sam} to generate precise, fine-grained masks for the objects.

Crucially, the labels generated by the Open Vocabulary Parser are open and determined solely by the interpretation of OMG-LLaVA, independent of the original caption. Conversely, the Caption-Anchored Parser restricts its labels strictly to objects explicitly mentioned in the original caption. Consequently, the mask-label pairs produced by these two parsers represent fundamentally distinct sets. We therefore conduct experiments on MaLSF to evaluate the mask-label pairs obtained from each parser independently.

\subsection{Global and Local Semantic Extraction}
Inspired by \cite{wang2024exploiting, zhang2025asap}, we propose a dual-path semantic extraction framework comprising three components: (1) mask-label pairs extraction, (2) textual feature encoding, and (3) multi-scale visual feature extraction.

\noindent \textbf{Mask-Label Pair Extraction.} Our local semantic parser (implementation details in Section.\ref{pipeline}) processes image $\mathcal{I}$ to generate $N$ mask-text pairs $\{(\mathbf{M}_i, \mathbf{L}_i)\}_{i=1}^N$, where binary mask $\mathbf{M}_i \in \{0,1\}^{H \times W}$ identifies semantic regions and $\mathbf{L}_i$ provides textual descriptions. The spatial-textual correspondence enables fine-grained cross-modal alignment.

\noindent \textbf{Textual Feature Encoding.} Shared text encoder $E_t$ processes both media caption $T$ and region labels $\{\mathbf{L}_i\}$. Let $\mathbf{l}_{\text{cap}}$ and $\{\mathbf{l}_i\},i=1,2,...,N$ represent the text features of the caption and region labels, respectively. All text features share latent space through weight sharing, with [CLS] tokens capturing global semantics.

\label{MVFE}
\noindent \textbf{Multi-scale Visual Feature Extraction.} Elements in different regions of an image have different visual features. To represent their overall visual semantics, we add a [CLS] token to the global and local visual features. Transformer-Based Visual encoder $E_v$ encodes the original image features as $\mathbf{v} \in \mathbb{R}^{H \times W \times C}$. 

For the global feature, we define it as $\mathbf{V}_{\text{img}} = [\mathbf{v}^{\text{cls}}_{\text{img}}, \mathbf{v}]$, where $\mathbf{v}^{\text{cls}}_{\text{img}}$  is the result of $\mathbf{v}$ passing through global average pooling. For the $i$-th local feature with the mask $\mathbf{M}_i$, we define it as $\mathbf{V}_i =  [\mathbf{v}^{\text{cls}}_i, \mathbf{v}_i^{\text{mask}}]$:
\begin{small}
\begin{align}
    \mathbf{v}_i^{\text{mask}} = \mathbf{M}_i \odot \mathbf{v},& \quad \mathbf{v}^{\text{cls}}_i = \frac{1}{|\Omega_i|}\sum\nolimits_{(h,w)\in \Omega_i}\mathbf{v}_{h,w}, 
\end{align}
\end{small}
where $\Omega_i = \{(x,y)|\mathbf{M}_i(x,y) = 1\}$, $\odot$ indicates element-wise multiplication. From this, we obtain the global semantic $\mathbf{V}_{\text{img}}$ and local semantic $\{\mathbf{V}_i\}$ of the visual features.

\subsection{Bidirectional Cross-modal Verification}
Let $\{\mathbf{V}_{img}, \mathbf{V}_1, ..., \mathbf{V}_N\}$ and $\{\mathbf{l}_{\text{cap}}, \mathbf{l}_1, ..., \mathbf{l}_N\}$ denote the encoded visual and textual features from previous processing, respectively. Here $\mathbf{V}_{img}$ and $\mathbf{l}_{cap}$ represent the original media content, while the remaining capture local semantics. Based on these, we propose a gated cross-attention mechanism to improve the noise robustness of verification.

Not all mask-label pairs are useful for manipulation detection. To automatically select informative local semantics, we first compute relevance weights through dual alignment:
\begin{small}
\begin{equation}
\begin{aligned}
   w_i^v & = \sigma\left(\phi_l(\mathbf{l}_{\text{cap}}^{cls})^\top \phi_v(\mathbf{v}_i^{cls})\right), \\
   \quad w_j^l & = \sigma\left(\phi_v(\mathbf{v}_{\text{img}}^{cls})^\top \phi_l(\mathbf{l}_j^{cls})\right) ,
\end{aligned}
\end{equation}
\end{small}
where $\phi_v(\cdot)$ and $\phi_l(\cdot)$ are learnable linear projections. $\sigma$ denotes the sigmoid function. $\mathbf{v}^{cls}$ and $\mathbf{l}^{cls}$ represent [CLS] tokens from visual/textual features, respectively. These weights are used to measure the importance of each visual and textual local semantics.

Then we employ two transformer encoder based verification streams ($\mathcal{T}_V$ and $\mathcal{T}_L$) to detect inconsistencies.

\noindent \textbf{Image-as-Query Verification}. $\mathbf{V}_{img}$ queries $\{\mathbf{l}_{\text{cap}}, \mathbf{l}_j\}$ to detect contradictions between global image and text labels. $\mathbf{F}_{\text{img}}^k$ denotes the consistency between global image and $k$-th label text.
\begin{small}
\begin{equation}
\{\mathbf{F}_{\text{img}}^{\text{cap}}, \mathbf{F}_{\text{img}}^1,...,\mathbf{F}_{\text{img}}^N\} = \mathcal{T}_V\left(\mathbf{V}_{\text{img}}, [\mathbf{l}_{\text{cap}}, \{w_j^l\mathbf{l}_j\}]\right).
\end{equation}
\end{small}

\noindent \textbf{Text-as-Query Verification}. $\mathbf{l}_{\text{cap}}$ examines $\{\mathbf{V}_{Img}, \mathbf{V}_i\}$ to identify mismatches between the caption and masked regions. $\mathbf{F}_{\text{cap}}^k$ denotes the consistency between the caption and $k$-th visual region.
\begin{small}
    \begin{equation} 
\{\mathbf{F}_{\text{cap}}^{\text{img}}, \mathbf{F}_{\text{cap}}^1,...,\mathbf{F}_{\text{cap}}^N\} = \mathcal{T}_L\left(\mathbf{l}_{\text{cap}}, [\mathbf{V}_{\text{img}}, \{w_i^v\mathbf{V}_i\}]\right).
\end{equation}
\end{small}

This bidirectional design forces the model to resolve three-level conflicts.
First, it addresses global inconsistency, $\mathbf{F}_{\text{img}}^{\text{cap}}$ and $\mathbf{F}_{\text{cap}}^{\text{img}}$ reveal overall image-text mismatches. Second, it detects local inconsistency through discrepancies in $\{\mathbf{F}_{\text{img}}^k\}$ or $\{\mathbf{F}_{\text{cap}}^k\}$ , which indicate manipulated regions and tokens. Finally, it highlights cross-modal inconsistency by contrasting the image-driven and text-driven verification paths to expose various manipulations.

\subsection{Hierarchical Semantic Aggregation}

\begin{figure}[t]
  \centering
  \includegraphics[width=\linewidth]{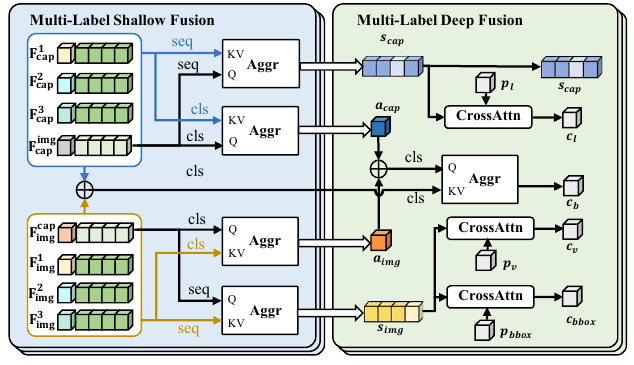}
  \caption{Architecture of Hierarchical Semantic Aggregation. All the verification features are put through multi-label shallow fusion and multi-label deep fusion, respectively, to get the final features input to the linear header.}
  \label{fig: HSA}
  \vspace{-0.3cm}
\end{figure}

The verification outputs $\{\mathbf{F}_{\text{img}}^{\text{cap}}, \{\mathbf{F}_{\text{img}}^k\}\}$ and $\{\mathbf{F}_{\text{cap}}^{\text{img}}, \{\mathbf{F}_{\text{cap}}^k\}\}$ require hierarchical aggregation to synthesize manipulation evidence. Our dual-stream architecture implements this through shallow and deep fusion stages as shown in Fig. \ref{fig: HSA}.

Similar to the preliminary extracted features in Sec.\ref{MVFE}, all verification outputs consist of [CLS] token and sequence tokens:
\begin{small}
\begin{equation}
    \begin{aligned}
        \mathbf{F}_{\text{img}}^{\text{cap}} & = [f_{\text{img}}^{\text{cls}},f_{\text{img}}^{\text{seq}}],\mathbf{F}_{\text{img}}^{k} = [f_{\text{img,k}}^{cls},f_{\text{img,k}}^{\text{seq}}], \\
        \mathbf{F}_{\text{cap}}^{\text{img}} & = [f_{\text{cap}}^{\text{cls}},f_{\text{cap}}^{\text{seq}}],\mathbf{F}_{\text{cap}}^{k} = [f_{\text{cap,k}}^{cls},f_{\text{cap,k}}^{\text{seq}}].
    \end{aligned}
\end{equation}
\end{small}

\noindent \textbf{Multi-Label Shallow Fusion.} For visual and text stream features, we decouple [CLS] tokens from sequence tokens. Our multi-label aggregation module is defined as:
\begin{small}
\begin{equation}
    \text{Aggr}(Q,KV) =Q + \text{LN}(\text{CrossAttn}(Q,KV,KV)),
\end{equation}
\end{small}
where the $\text{LN}$ means layer normalization. The $\text{CrossAttn}(\cdot)$ denotes the cross attention mechanism \cite{vaswani2017attention}. $Q$ and $KV$ are a query vector and a key-value vector, respectively. The aggregation for visual and text streams is then performed through:
\begin{small}
\begin{equation}
\begin{aligned}
a_{\text{img}} &= \text{Aggr}_{\text{img},\text{cls}}\left(f_{\text{img}}^{\text{cls}}, \text{Cat}(f_{\text{img}}^{\text{cls}},\{f_{\text{img},k}^{\text{cls}}\})\right), \\
s_{\text{img}} &= \text{Aggr}_{\text{img},\text{seq}}\left(f_{\text{img}}^{\text{seq}}, \text{Cat}(f_{\text{img}}^{\text{seq}},\{f_{\text{img},k}^{\text{seq}}\})\right),\\
a_{\text{cap}} &= \text{Aggr}_{\text{cap},\text{cls}}\left(f_{\text{cap}}^{\text{cls}}, \text{Cat}(f_{\text{cap}}^{\text{cls}},\{f_{\text{cap},k}^{\text{cls}}\})\right), \\
s_{\text{cap}} &= \text{Aggr}_{\text{cap},\text{seq}}\left(f_{\text{cap}}^{\text{seq}}, \text{Cat}(f_{\text{cap}}^{\text{seq}},\{f_{\text{cap},k}^{\text{seq}}\})\right),
\end{aligned}
\end{equation}
\end{small}
where $\text{Cat}(\cdot)$ denotes concatenation along the token dimension. $a_{\text{img}}$ and $a_{\text{cap}}$ (aggregated [CLS]) capture global consensus by attending to semantic centers, while $s_{\text{img}}$ and $s_{\text{cap}}$ (aggregated sequences) preserves fine-grained contextual relationships. This prevents information dilution between high-level semantics and low-level details.

\noindent \textbf{Multi-Label Deep Fusion.}
Consider that we have four types of tasks: true/false classification, manipulation type classification, image manipulation grounding, and text manipulation grounding. Therefore, we need to further fuse features for these tasks.

For the manipulation type classification task, we decouple the original four binary classifications into text-manipulation classification and image-manipulation classification, and used two dual features aggregated by querying for that task, respectively.
Let $p_v, p_l \in \mathbb{R}^D$ be two aggregate tokens that can be learned, then we can obtain:
\begin{small}
\begin{equation}
\begin{aligned}
c_v &= \text{CrossAttn}_{v}\left(p_v, s_{\text{img}}, s_{\text{img}}\right),  \\
c_l &= \text{CrossAttn}_{l}\left(p_l, s_{\text{cap}}, s_{\text{cap}}\right)  ,
\end{aligned}
\end{equation}
\end{small}
\noindent where $c_v$ and $c_l$ are the features used for image and text manipulation classification, respectively.

For the binary classification task, we need to use the cls features $a_{\text{img}}, a_{\text{cap}}$ obtained in shallow fusion. We use the added cls token to aggregate the verification outputs again:
\begin{footnotesize}
\begin{equation}
c_{\text{b}} = \text{Aggr}\left(a_{\text{img}} \oplus a_{\text{cap}}, \text{Cat}((f_{\text{img}}^{\text{cls}},\{f_{\text{img},k}^{\text{cls}}\}) \oplus (f_{\text{cap}}^{\text{cls}},\{f_{\text{cap},k}^{\text{cls}}\}))\right),
\label{eq:binary cls}
\end{equation}
\end{footnotesize}
where $\oplus$ denotes element-wise addition. 


For image manipulation grounding, we use $p_{\text{bbox}} \in \mathbb{R}^D$ to aggregate the visual feature $s_{\text{img}}$:
\begin{small}
\begin{equation}
    c_{\text{bbox}} = \text{CrossAttn}_{\text{bbox}}\left(p_{\text{bbox}}, s_{\text{img}}, s_{\text{img}}\right).
\end{equation}
\end{small}
Then we can get the bounding box by inputting $c_{\text{bbox}} \in \mathbb{R}^{D} $ to the linear classification header $H_{ig}: \mathbb{R}^{D} \to \mathbb{R}^{4} $. 


For text manipulation grounding, we need to determine the index of the manipulated token. We can get the prediction at each position by inputting $s_{\text{cap}} \in \mathbb{R}^{L \times D}$ to the linear classification header $H_{tg}: \mathbb{R}^{L \times D} \to \mathbb{R}^{L\times 2} $. 

We have thus obtained the tasks specific features $c_v,c_l,c_b,c_{\text{bbox}}$ and $s_{\text{cap}}$ decoupled by hierarchical aggregation. These features are used to calculate the losses after going through the corresponding classification head.

\subsection{Loss Function}
We calculate the loss for each task using the same method as \cite{liu2024unified,shao2023detecting}. Except loss $\mathcal{L}_{ig}$ for the image grounding task, which was calculated based on the L1 loss and GIoU loss \cite{rezatofighi2019generalized} of the bbox, the losses for the other three tasks ($\mathcal{L}_{bcls},\mathcal{L}_{mcls},\mathcal{L}_{tg}$) were obtained using cross-entropy loss.\footnote{A complete definition of each type of loss is provided in the Appendix.} We combine the above losses to obtain the final loss function, where the $\alpha$, $\beta$, and $\gamma$ are hyperparameters that control the importance:
\begin{small}
\begin{equation}
    \mathcal{L} = \mathcal{L}_{bcls} + \alpha \mathcal{L}_{mcls} + \beta \mathcal{L}_{ig} + \gamma \mathcal{L}_{tg}.
\end{equation}
\end{small}

\vspace{-0.3cm}
\section{Experiments}
We will conduct comprehensive experiments on the tasks of DGM4 and MFND. For MFND, we train our model with only the binary classification loss $\mathcal{L}_{bcls}$. More information on datasets and evaluation metrics can be found in the Appendix.

\begin{table*}[t]
\setlength\tabcolsep{5.65pt}
\centering
\small
\begin{tabular*}{0.99 \textwidth}{@{\extracolsep{\fill}}{l}c*{11}{c}|{c}@{}}
\toprule
Categories & 
\multicolumn{3}{c}{Binary Cls(\%)} & 
\multicolumn{3}{c}{Multi-Label Cls(\%)} & 
\multicolumn{3}{c}{Image Grounding(\%)} & 
\multicolumn{3}{c|}{Text Grounding(\%)} & \multirow{2}{*}{$\Delta$Avg}\\
\cmidrule(lr){2-4} \cmidrule(lr){5-7} \cmidrule(lr){8-10} \cmidrule(lr){11-13}
Methods & AUC & EER$\downarrow$ & ACC & OF1 & CF1 & mAP & $\text{IoU}_{\text{mean}}$ & $\text{IoU}_{\text{50}}$ & $\text{IoU}_{\text{75}}$ & Prec & Recall & F1 \\
\midrule
CLIP~\cite{radford2021learning}  & 83.22 & 24.61 & 76.40 & 66.00 & 59.52 & 62.31 & 49.51 & 50.03 & 38.79 & 58.12 & 22.11 & 32.03 & --\\
ViLT~\cite{kim2021vilt} &   85.16 & 22.88 & 78.38 & 72.37 & 66.14 & 66.00 & 59.32 & 65.18 & 48.10 & 66.48 & 49.88 & 57.00 & --\\
HAMMER~\cite{shao2023detecting} &  93.19 & 14.10 & 86.39 & 80.37 & 79.37 & 86.22 & 76.45 & 83.75 & 76.06 & 75.01 & 68.02 & 71.35 &  0\\
VIKI~\cite{li2024towards} &  93.51 & 13.87 & 86.67 & 80.10 & 81.07 & 86.58 & 76.51 & 83.95 & 75.77 & 77.79 & 66.06 & 72.44 & +0.38\\
HAMMER++~\cite{shao2024detecting} & 93.33 & 14.06 & 86.66 & 80.71 & 79.73 & 86.41 & 76.46 & 83.77 & 76.03 & 73.05 & 72.14 & 72.59 & +0.40 \\
UFAFormer~\cite{liu2024unified} & 93.81 & 13.60 & 86.80 & 81.48 & 80.31 & 87.85 & 78.33 & 85.39 & 79.20 & 73.35 & 70.73 & 72.02 & +1.13\\
ASAP~\cite{zhang2025asap}& 94.38 & 12.73 & 87.71 & 82.89 & 81.72 & 88.53 & 77.35 & 84.75 & 76.54 & 79.39 & 73.86 & 76.52 & +2.40\\
EMSF~\cite{wang2024exploiting} & 95.11 & 11.36 & 88.75 & 84.38 & 83.60 & \textbf{91.42} & 80.83 & 88.35 & 80.39 & 76.51 & 70.61 & 73.44 & +3.33\\
\rowcolor{gray!20} \textbf{$\text{MaLSF}^\bigstar$(Ours)} & \underline{95.56} & \underline{10.92} & \underline{89.33} & \underline{84.92} & \underline{83.84} & \underline{90.76} & \textbf{82.47} & \underline{88.83} & \textbf{84.62} & \underline{81.59} & \underline{72.68} & \underline{76.88} & \underline{+4.87}\\
\rowcolor{gray!20} \textbf{$\text{MaLSF}^\Diamond$(Ours)} &  \textbf{95.60} & \textbf{10.89} & \textbf{89.37} & \textbf{85.07} & \textbf{83.93} & 90.46 & \underline{82.37} & \textbf{88.95} & \underline{84.48} & \textbf{81.62} & \textbf{73.22} & \textbf{77.19} &\textbf{+4.94}\\
\bottomrule
\end{tabular*}
\caption{Comparison with state-of-the-art methods for multimodal media manipulation detection and grounding on DGM4. Bold and underlined text indicates the best and second-best performance metrics, respectively. $\text{MaLSF}^\bigstar$ and $\text{MaLSF}^\Diamond$ refer to the mask-label pairs obtained using the Caption-Anchored Parser and Open Vocabulary Parser, respectively. All metrics are reported in percentages (\%).
}
\label{tab:comparison}
\end{table*}

\begin{table*}[t]
\setlength\tabcolsep{6pt}
\centering
\small
\begin{tabularx}{0.99 \textwidth}{lXcc*{6}{c}}
\toprule
\multirow{2}{*}{} & \multirow{2}{*}{Method} & \multirow{2}{*}{Reference} & \multirow{2}{*}{Accuracy(\%)} & 
\multicolumn{3}{c}{Fake News} & 
\multicolumn{3}{c}{Real News} \\
\cmidrule(lr){5-7} \cmidrule(lr){8-10}
 & & & & Precision(\%) & Recall(\%) & F1(\%) & Precision(\%) & Recall(\%) & F1(\%) \\
\midrule
\multirow{5}{*}{\rotatebox{90}{\textbf{Weibo21}}} 
& CAFE \cite{chen2022cross} & WWW'22 & 88.2 & 85.7 & 91.5 & 88.5 & 90.7 & 84.4 & 87.6 \\
&  BMR \cite{10.1609/aaai.v37i4.25670} & AAAI'23 & 92.9 & 90.8 & 94.7 & 92.7 & 94.6 & 90.6 & 92.5 \\
& FND-CLIP \cite{zhou2023multimodal} & ICME'23 & 94.3 & 93.5 & 94.5 & 94.0 & 95.0 & 94.2 & 94.6 \\
& DAMMFND \cite{lu2025dammfnd} & AAAI'25 & 94.7 & -- & -- & 94.8 & -- & -- & 94.7 \\
& MMDFND \cite{tong2024mmdfnd}  & MM'24 & 93.9 & -- & -- & 94.0 & -- & -- & 93.9 \\
\rowcolor{gray!20}& \textbf{$\text{MaLSF}^\Diamond$(Ours)} & -- & 95.1 & 94.9 & 95.5 & 95.2 & 95.4 & 94.8 & 95.1 \\
\rowcolor{gray!20}& \textbf{$\text{MaLSF}^\bigstar$(Ours)} & -- & \textbf{95.5} & \textbf{95.2} & \textbf{95.8} & \textbf{95.5} & \textbf{95.7} & \textbf{95.1} & \textbf{95.4} \\

\hline
\multirow{6}{*}{\rotatebox{90}{\textbf{Weibo17}}}
& FND-CLIP \cite{zhou2023multimodal} & ICME'23 & 90.7 & 91.4 & 90.1 & 90.8 & 91.4 & 90.1 & 90.7 \\
& BMR \cite{10.1609/aaai.v37i4.25670} & AAAI'23 & 91.8 & 88.2 & \textbf{94.8}& 91.4 & 94.2 & 87.9 & 90.4 \\
& COOLANT \cite{wang2023cross} & MM'23 & 92.3 & 92.7 & 92.3 & 92.5 & 91.9 & 92.2 & 92.0 \\
&  MIMoE-FND \cite{liu2025modality} & WWW'25 & 92.8 & 94.2 & 91.3 & 92.8 & 91.3 & 94.2 & 92.7 \\
& RaCMC \cite{yu2025racmc} & AAAI'25  & 91.5 & 91.0 & 92.4 & 91.7 & 92.1 & 90.6 & 91.4 \\
& FSRU \cite{lao2024frequency} & AAAI'24 & 90.1 & 92.2 & 89.2 & 90.6 & 87.9 & 91.3 & 89.5 \\
\rowcolor{gray!20}& \textbf{$\text{MaLSF}^\Diamond$(Ours)} & -- & 93.4 & 92.0 & 94.5 & \textbf{93.3} & \textbf{94.7} & 92.3 & 93.5 \\
\rowcolor{gray!20}& \textbf{$\text{MaLSF}^\bigstar$(Ours)} & -- & \textbf{93.5} & \textbf{94.8} & 91.7 & 93.2 & 92.4 & \textbf{95.3} & \textbf{93.8} \\
\bottomrule
\end{tabularx}
\caption{Comparison with state-of-the-art methods for multimodal fake news detection on Weibo17 and Weibo21.
Bold and underlined text indicates the best and second-best performance metrics, respectively. $\text{MaLSF}^\bigstar$ and $\text{MaLSF}^\Diamond$ refer to the mask-label pairs obtained Caption-Anchored Parser and Open Vocabulary Parser, respectively.}
\label{tab: mfnd}
\vspace{-0.2cm}
\end{table*}

\subsection{Implementation Details}
\label{Implementation Details}
The visual encoder is implemented by Swin-B \cite{Liu_2021_ICCV}, and the text encoder is implemented by a six-layered BERT encoder\cite{devlin2019bert}. The modality verification module is constructed using 6 BERT encoder layers with cross attention. We load the X-VLM \cite{zeng2021multi} as the pretrain weights. The binary classifier, fine-grained classifier, and bbox detector are all implemented using multilayer perceptron layers. We utilize the AdamW \cite{loshchilov2019decoupledweightdecayregularization} optimizer with a weight decay of 0.02. During the first 10 epochs, the learning rate is warmed up to 1e-5 for the text branch and 5e-5 for the image branch and then decays to 6e-7 using a cosine schedule during 60 epochs of training. Our hyperparameters are set to: $\alpha=1.5$, $\beta=0.1$, $\gamma=1$. We train our model on 8 RTX 4090 GPUs for about 8 hours. For DGM4, the length of the text content is padded or truncated to 50 tokens, while the images are resized to 256x256 pixels.  For MFND, we set the maximum length of the text content to 128, which is even less than that of \cite{liu2025modality,zhou2023multimodal,wang2023cross,10.1609/aaai.v37i4.25670}. Other settings are identical to those of state-of-the-art methods, ensuring a fair comparison of results.

\vspace{-0.1cm}
\subsection{Datasets}

\noindent\textbf{DGM4.} For media manipulation detection and grounding, we used the DGM4 dataset \cite{shao2023detecting}. This is the only dataset currently available for this task. DGM4 consists of 230k image-text pairs with more than 77k original pairs and 152k processed pairs. In the DGM4 dataset, image manipulation includes face swapping (FS) and facial attribute manipulation (FA), while text manipulation includes text swapping (TS) and text attribute manipulation (TA).

\noindent \textbf{Weibo17,Weibo21.} 
For multimodal fake news detection, we choose the Weibo17 \cite{10.1145/3123266.3123454} dataset and the Weibo21 \cite{10.1145/3459637.3482139} dataset. Both Weibo17 and Weibo21 are Chinese datasets containing image and text pairs collected from the social media platform Weibo. We keep the same data split scheme as \cite{chen2022cross,wang2023cross} for Weibo17 and keep the same train-test split at a ratio of 9:1 of \cite{wang2023cross,10.1609/aaai.v37i4.25670}.

\subsection{Performance Comparison}

\noindent \textbf{Comparison with the state-of-the-art methods of DGM4.} 
We compare MaLSF with state-of-the-art methods CLIP \cite{radford2021learning}, ViLT \cite{kim2021vilt}, HAMMER \cite{shao2023detecting}, HAMMER++ \cite{shao2024detecting}, VIKI \cite{li2024towards}, UFAFormer \cite{liu2024unified}, ASAP \cite{zhang2025asap}, EMSF \cite{wang2024exploiting} on DGM4 across four tasks: binary classification, multi-label classification, image grounding, and text grounding as shown in Tab. \ref{tab:comparison}. $\text{MaLSF}^\Diamond$ and $\text{MaLSF}^\bigstar$ use mask-label pairs from Caption-Anchored Parser and Open Vocabulary Parser, respectively. For multi-label classification, it yields the best OF1 and CF1, with mAP slightly below EMSF. Nevertheless, MaLSF achieves average performance improvements of 4.87\% and 4.94\% across all tasks, far exceeding other methods. Overall, MaLSF outperforms prior methods on 11 of 12 metrics, highlighting its robustness. Notably, pairs from the Open Vocabulary Parser better support text manipulation grounding for its open domain labels.

\noindent \textbf{Comparison with the state-of-the-art methods of MFND.} 
To evaluate the effectiveness of MaLSF on multimodal fake news detection (MFND), we compared it with state-of-the-art methods on Weibo21 and Weibo17 (Tab. \ref{tab: mfnd}). $\text{MaLSF}^\Diamond$ and $\text{MaLSF}^\bigstar$ achieve state-of-the-art results on both datasets. On Weibo21, MaLSF-OV reaches 95.5\% accuracy and F1 scores, surpassing DAMMFND \cite{lu2025dammfnd} (94.7\%). On Weibo17, it achieves 93.5\%, outperforming MIMoE-FND \cite{liu2025modality} (92.8\%). The consistent gains highlight the robustness of our multimodal fusion, with $\text{MaLSF}^\Diamond$ slightly outperforming $\text{MaLSF}^\bigstar$, showing better adaptability to diverse multimodal patterns.

\begin{table}[t]
\centering
\small
\begin{tabular}{@{}ccccccc@{}}
\toprule
\multicolumn{3}{c}{Components} & 
\multicolumn{4}{c}{Performance Metrics(\%)} \\
\cmidrule(lr){1-3} \cmidrule(lr){4-7}
Label & Mask & ML Pair & ACC  & CF1  & $\text{IoU}_{\text{mean}}$  & F1   \\
\midrule
  &  &  & 88.93 & 82.91 & 76.12 &  75.59 \\
\checkmark &  &  & \underline{89.26} & 83.43 & 80.78 &  \underline{76.69}   \\
 & \checkmark &  & 89.24 & \underline{83.60} & \underline{82.02} & 75.90   \\
\rowcolor{gray!20} &  & \checkmark & \textbf{89.37} & \textbf{83.93} & \textbf{82.37} &  \textbf{77.19}  \\
\bottomrule
\end{tabular}
\caption{Ablation study of different modal local semantics in the proposed method. “Label" means only text labels are used, “Mask" means only image masks are used, and “ML pair" means the matching mask-label pairs are used. }
\label{tab:ablation_modal}
\vspace{-0.2cm}
\end{table}

\begin{table}[t]
\centering
\small
\begin{tabular}{@{}cccc*{4}c@{}}
\toprule
\multicolumn{4}{c}{Components} & 
\multicolumn{4}{c}{Performance Metrics(\%)} \\
\cmidrule(lr){1-4} \cmidrule(lr){5-8}
I.B & T.B & T.V & I.V & ACC  & CF1  & $\text{IoU}_{\text{mean}}$  & F1  \\
\midrule
& \checkmark &  &\checkmark & 88.03 & 81.65 & 79.97 & --  \\
\checkmark & & \checkmark & & \underline{88.95} & 83.11 & 79.77 & 76.28\\
\checkmark & \checkmark &  \checkmark & & 88.87 & \underline{83.46} & \underline{81.42} & \underline{76.53}\\
\rowcolor{gray!20} \checkmark & \checkmark & \checkmark & \checkmark & \textbf{89.37} & \textbf{83.93} & \textbf{82.37} & \textbf{77.19} \\
\bottomrule
\end{tabular}
\caption{Ablation study of BCV modules. “I.B" and “T.B" denote Image Branch and Text Branch. “T.V" and “I.V" indicate text and image verification, which means using text or image as a query.}
\label{tab:ablation2}
\vspace{-0.3cm}
\end{table}

\begin{table}[h]
\centering
\small
\begin{tabular}{@{}ccccc*{3}c@{}}
\toprule
\multicolumn{4}{c}{Components} & 
\multicolumn{4}{c}{Performance Metrics(\%)} \\
\cmidrule(lr){1-4} \cmidrule(lr){5-8}
ML & GW & MSF & MDF & ACC  & CF1  & $\text{IoU}_{\text{mean}}$  & F1  \\
\midrule
 &  &  &  & 88.90 & 83.38 & 81.74 & 75.87  \\
\checkmark &  &  &  & 87.81 & 81.30 & 80.77 & 75.60\\
\checkmark & \checkmark & & & 87.97 & 81.72 & 79.55 & 75.59 \\
\checkmark &  & \checkmark & \checkmark & \underline{89.18} & \underline{83.65}& 81.35 & \underline{76.97}   \\
\checkmark & \checkmark & \checkmark &  & 89.08 & 83.45 & \underline{81.66} & 76.52 \\
\rowcolor{gray!20} \checkmark & \checkmark & \checkmark & \checkmark & \textbf{89.37} & \textbf{83.93} & \textbf{82.37} & \textbf{77.19}  \\
\bottomrule
\end{tabular}
\caption{Ablation study of the proposed modules. “ML" denotes mask-label pairs. “GW" denotes the gate weight module. “MSF" and “MDF" denote multi-label shallow fusion and multi-label deep fusion, respectively.}
\label{tab:ablation1}
\vspace{-0.3cm}
\end{table}


\subsection{Ablation Study}

\noindent \textbf{Ablation study of two modalities on DGM4.}
To verify that our model is aware of the association between labels and masks, we performed ablation experiments based on the Open Vocabulary parser as shown in Tab. \ref{tab:ablation_modal}. We report representative metrics (ACC, CF1, IOU, F1) for the four tasks. We can see that the overall performance of the model is not impressive when no local semantics are introduced. When only the labels of the text modality are introduced, we can see a slight improvement in all the metrics except the performance of text grounding. When only the image modal mask is introduced, the performance in all aspects is further improved. Combining mask-label pairs produces the best results, demonstrating that MaLSF effectively leverages both modalities to establish local associations.

\begin{figure*}[h]
  \centering
  \includegraphics[width=\linewidth]{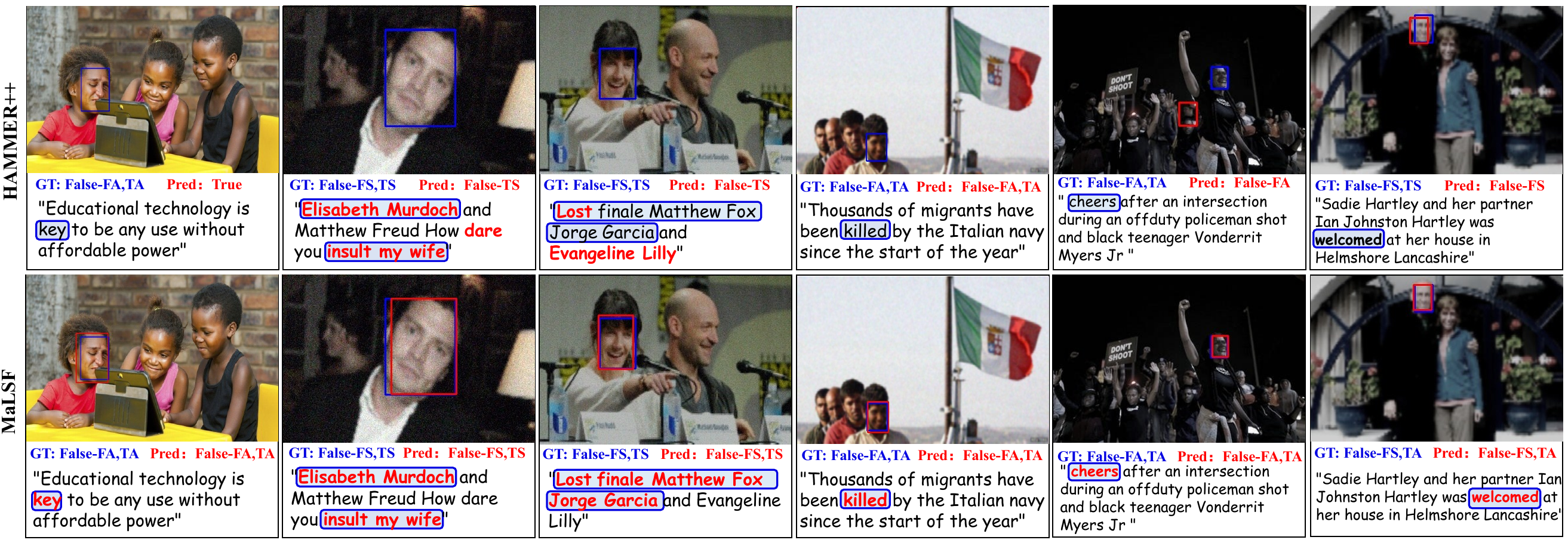}
  \caption{Qualitative analysis of results for HAMMER++ and MaLSF. The red bbox and text indicate the model prediction results. The blue text boxes and bounding boxes are ground truth. MaLSF is superior in both detection and grounding tasks.}
  \label{fig: vis_comp_double}
  \vspace{-0.2cm}
\end{figure*}


{

\noindent \textbf{Ablation study of proposed modules on DGM4.}
Tab. \ref{tab:ablation2} highlights the contributions of our BCV module. Image Branch and Text Branch are the encoding layers with self-attention. “T.V" and “I.V" indicate whether using text or image features as a query. We can see that the effectiveness of text unidirectional verification (“I.B" plus “T.V") and image unidirectional verification (“T.B" plus “I.V") are inferior to that of our bidirectional verification.
To verify the effectiveness of our proposed other modules, we conduct ablation experiments as shown in Tab. \ref{tab:ablation1}. Without multi-label shallow (MSF) and deep (MDF) fusion in HSA, semantics are fused by addition, leading to degraded performance even with mask-label pairs (ML) and gate weights (GW) in BCV. The improvement after adding our MSF and MDF sequentially is significant. Additionally, comparing the experimental results without GW, we find GW contributes significantly to improving the effect of image manipulation grounding. Overall, these results verify the effectiveness of our BCV and HSA modules.

\subsection{Visualization and Analysis of Results}

\noindent \textbf{Qualitative analysis of results on DGM4.}
Fig. \ref{fig: vis_comp_double} compares the grounding and detection results of our MaLSF with HAMMER++ \cite{shao2024detecting}. The red bbox and text indicate the grounding prediction results. The blue ones indicate the ground truth. HAMMER++ occasionally misclassifies manipulations, leading to incorrect or missed detections. Even when classifying correctly, HAMMER++ often struggles with grounding or precise localization. In contrast, MaLSF achieves more accurate classification and significantly better manipulation detection and grounding. The outstanding qualitative results are consistent with the excellent performance of MaLSF on evaluation metrics.


\begin{figure}[t]
  \centering
  \includegraphics[width=\linewidth]{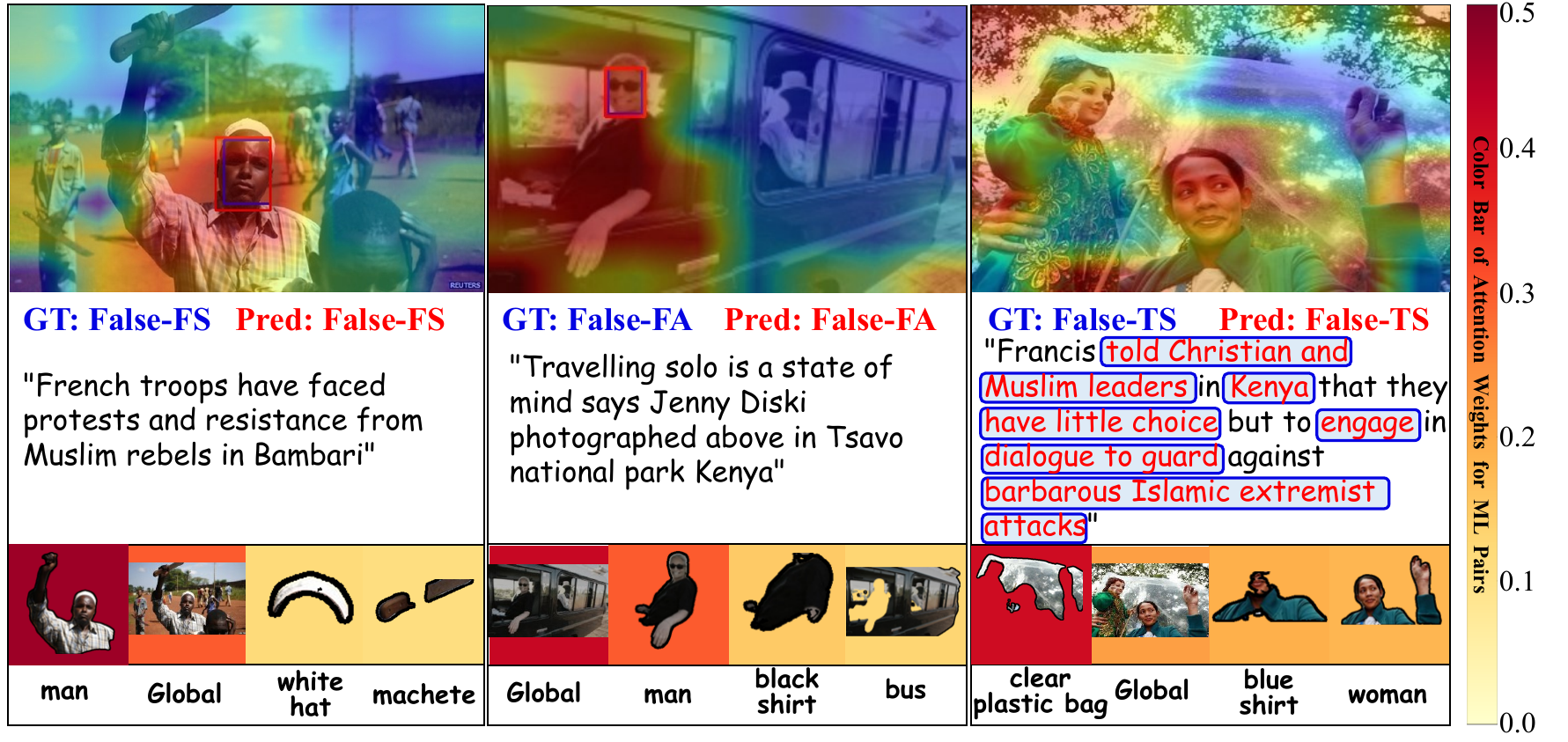}
  \caption{Visualization of spatial attention of image features and the weights of local semantics. The mapping of local semantic weights to colors is shown as the color bar.}
  \label{fig: vis_label}
  \vspace{-0.5cm}
\end{figure}

\noindent \textbf{Qualitative analysis of local semantic fusion on DGM4.}
To analyze fusion effect across local semantics of MaLSF, we visualize the fusion weights (i.e., attention weights in Equ.\ref{eq:binary cls}) of mask-label pairs from the Open Vocabulary Parser. Higher weights indicate greater semantic importance. As shown in Fig. \ref{fig: vis_label}, ‘Global’ indicates the global semantics of the original image and text. For images with face manipulation (FS, FA), mask-label pairs covering manipulated facial regions have high weights. For text-swapped (TS) content, the most conflicting pair (“a clear plastic bag") attains the highest weight. In addition, spatial attention visualization on the image confirms that MaLSF focuses precisely on manipulated facial areas. From the weights of the mask-label pairs, we can see the criteria used by the model to make its judgments. This demonstrates the interpretability of our MaLSF.

\subsection{Model Efficiency Comparison.}
We compare the number of trainable parameters and inference speed of our model with those of the currently open-source model HAMMER++\cite{shao2024detecting}, ASAP\cite{zhang2025asap}, and EMSF\cite{wang2024exploiting}. As shown in Tab. \ref{tab:efficiency}, our number of trainable parameters is the lowest compared to the other two state-of-the-art methods. And we are also able to achieve an inference speed of 22.9 per second. Though this is slightly slower than the 31.1 frames-per-second (FPS) of HAMMER++, MaLSF shows a substantial improvement in performance with a competitive inference speed.

\begin{table}[h]
\centering
\small
\begin{tabular}{@{} p{2cm} llll@{}}
\toprule
Method & Params & FPS & Avg. Pref.(\%) \\ \hline
HAMMER++\cite{shao2024detecting} & 441M & \textbf{31.1} &  72.23\\
ASAP\cite{zhang2025asap} & 441M & \textbf{31.1} & 74.24 \\
EMSF \cite{wang2024exploiting} & 328M &  -- & 75.17 \\
\textbf{MaLSF(ours)} & \textbf{319M} &  22.9 & \textbf{76.88}\\
\bottomrule
\end{tabular}
\caption{Comparison of the efficiency of the MaLSF with HAMMER++\cite{shao2024detecting}, ASAP\cite{zhang2025asap} and EMSF\cite{wang2024exploiting}.}
\label{tab:efficiency}
\vspace{-0.5cm}
\end{table}

\section{Conclusion}
The proposed MaLSF framework tackles the challenge of fine-grained semantic alignment in multimodal media verification by introducing mask-aware local semantic fusion. Through its Bidirectional Cross-modal Verification (BCV) and Hierarchical Semantic Aggregation (HSA) modules, MaLSF achieves state-of-the-art performance in DGM4 and MFND. The framework explicitly models mask-label pairs to localize cross-modal inconsistencies, improving interpretability. This work advances multimodal verification by bridging pixel-level and textual semantics, providing a scalable solution against misinformation. Future research may extend the framework to video-text domains and explore self-supervised mask-label pair generation.

\section*{Acknowledgments}
This research was supported by the National Natural Science Foundation of China (No. U23B2060, No. 62495092).

{
    \small
    \bibliographystyle{ieeenat_fullname}
    \bibliography{main}
}

\clearpage
\setcounter{page}{1}
\maketitlesupplementary

\section{Loss Calculation}
\label{appendix:loss}
For the binary classification, the loss is obtained by calculating the cross-entropy loss between the classification head ($H_{\text{b}}: \mathbb{R}^{D} \to \mathbb{R}^2$) outputs and the ground truth labels  $y_{\text{b}}$.
\begin{align}
\mathcal{L}_{\text{bcls}} & = \mathcal{L}_{\text{ce}}(L_{\text{b}}(c_{\text{b}}),y_{\text{b}}), 
\end{align}
where $\mathcal{L}_{\text{ce}}$ denotes the cross-entropy loss.

For the manipulation type classification task, we input features $c_{\text{v}}$ and $c_{\text{l}}$ to classification heads $H_{\text{v}}: \mathbb{R}^{D} \to \mathbb{R}^2$ and $H_{\text{l}}: \mathbb{R}^{D} \to \mathbb{R}^2$, respectively, and calculate the loss as follows:
\begin{equation}
    \mathcal{L}_{\text{mcls}}  = \mathcal{L}_{\text{ce}}(\text{Cat}(H_{\text{v}}(c_{\text{v}}),H_{\text{l}}(c_{\text{l}})),y_{\text{m}}) ,
\end{equation}
where $y_{\text{m}}$ represents the truth labels for manipulation type classification and “Cat" indicates the concatenation operation.

For text manipulation grounding, we need to determine the index of the manipulated token. We can get the prediction at each position by inputting $s_{\text{cap}} \in \mathbb{R}^{L \times D}$ to the linear classification header $L_{tg}: \mathbb{R}^{L \times D} \to \mathbb{R}^{L\times 2} $. The loss for the text grounding is: 
\begin{equation}
    \mathcal{L}_{\text{tg}} = \mathcal{L}_{ce}(H_{tg}(s_{\text{cap}}),y_{tg}),
\end{equation}
where the $y_{tg} = \{y_i\}_{i=1}^{L}$ denotes whether the $i$-th token is manipulation or not.

For image manipulation grounding, we can get the bounding box by inputting $c_{\text{bbox}} \in \mathbb{R}^{D} $ to the linear classification header $H_{ig}: \mathbb{R}^{D} \to \mathbb{R}^{4} $. Then the loss is:
\begin{small}
\begin{equation}
    \mathcal{L}_{\text{ig}} = \mathcal{L}_{\text{L1}}(H_{ig}(c_{\text{bbox}})-y_{ig})+\mathcal{L}_{\text{GIoU}}(H_{ig}(c_{\text{bbox}})-y_{ig}),
\end{equation}
\end{small}
where the $y_{ig}$ represents the manipulated image grounding label. The $\mathcal{L}_{\text{L1}}$ and $\mathcal{L}_{\text{GIoU}}$ denote the L1 loss and GIoU loss \cite{rezatofighi2019generalized}.

We combine the above losses to obtain the final loss function, where the $\alpha$, $\beta$, and $\gamma$ are hyperparameters that control the importance:
\begin{equation}
    \mathcal{L} = \mathcal{L}_{bcls} + \alpha \mathcal{L}_{mcls} + \beta \mathcal{L}_{ig} + \gamma \mathcal{L}_{tg}.
\end{equation}
In our experiment, we used the following hyperparameters: $\alpha=1.5$, $\beta=0.1$, $\gamma = 1$.

\subsection{Implementation Details of Datasets}
\label{datasets}
Since DGM4 is the only dataset currently available for media manipulation detection and grounding, in order to more comprehensively validate the effectiveness of our approach, we also conducted experiments on the multimodal fake news detection datasets Weibo21 and Weibo17. DGM4 is an artificial dataset, and Weibo17 and Weibo21 are real datasets; the excellent results achieved on both of them also serve to further demonstrate the applicability of our model.

\noindent\textbf{DGM4.} For media manipulation detection and grounding, we used the DGM4 dataset\cite{shao2023detecting}. This is the only dataset currently available for this task. DGM4 consists of 230k image-text pairs with more than 77k original pairs and 152k processed pairs. The DGM4 dataset is constructed based on the VisualNews dataset [35], which was collected from several news agencies. In the DGM4 dataset, image manipulation includes face swapping (FS) and facial attribute manipulation (FA), while text manipulation includes text swapping (TS) and text attribute manipulation (TA).

\noindent \textbf{Weibo17,Weibo21.} 
For multimodal fake news detection, we choose the Weibo17 dataset and the Weibo21 dataset. Both Weibo17 and Weibo21 are Chinese datasets containing image and text pairs collected from the social media platform Weibo. 
The Weibo dataset collected by \cite{10.1145/3123266.3123454} contains 3749 fake news and 3783 real news for training, 1000 fake news and 996 real news for testing. In experiments, we follow the same steps in the work \cite{10.1145/3123266.3123454,wang2018eann} to remove the duplicated and low-quality images to ensure the quality of the entire dataset. We keep the same data split scheme as \cite{chen2022cross,wang2023cross}. 
The Weibo21 was created in 2021 by \cite{10.1145/3459637.3482139}, where more recent social posts were collected with 4640 real news and 4487 fake news in total. We keep the same train-test split at a ratio of 9:1 of \cite{wang2023cross,10.1609/aaai.v37i4.25670}.

\subsection{Evaluation Metrics}
\label{eva metrics}
\noindent \textbf{DGM4.} The evaluation metrics on DGM4 consist of a total of 12 metrics for the four tasks. For binary classification, we evaluate the Accuracy (ACC), Area Under the receiver operating characteristic curve (AUC), and equal error rate (EER). For manipulation classification, we evaluate mean mean F1 per class, (CF1) overall F1 (OF1), and mean average precision (MAP). For image manipulation grounding, we evaluate mean intersection over union (IoUmean), and IoU at thresholds of 0.5 (IoU50) and 0.75 (IoU75). For text manipulation grounding, we evaluate precision, recall, and F1 score.

\noindent \textbf{Weibo17,Weibo21.} The evaluation metrics of fake news detection include Accuracy on all samples. The metrics evaluated on real and fake news are Precision, recall and F1-score. 

\end{document}